\documentclass[10pt,twocolumn,letterpaper]{article}

\usepackage[pagenumbers]{cvpr} 

\usepackage{graphicx}
\usepackage{amsmath}
\usepackage{amssymb}
\usepackage{booktabs}
\usepackage[pagebackref,breaklinks,colorlinks]{hyperref}
\usepackage[capitalize]{cleveref}
\crefname{section}{Sec.}{Secs.}
\Crefname{section}{Section}{Sections}
\Crefname{table}{Table}{Tables}
\crefname{table}{Tab.}{Tabs.}

\def\confName{CVPR}
\def\confYear{2023}

\begin{document}

\title{3D reconstruction using Structure for Motion}

\author{Kshitij Karnawat\\
kshitij\\
{\tt\small kshitij@terpmail.umd.edu}
\and
Hritvik Choudhari\\
hac\\
{\tt\small hac@umd.edu}
\and
Abhimanyu Saxena\\
asaxena4\\
{\tt\small asaxena4@umd.edu}
\and
Mudit Singal\\
msingal\\
{\tt\small msingal@umd.edu}
\and
Raajith Gadam\\
raajithg\\
{\tt\small raajithg@umd.edu}
}
\maketitle

\begin{abstract}
 We are working towards 3D reconstruction of indoor spaces using a pair of HDR cameras in a stereo vision configuration mounted on an indoor mobile floor robot that captures various textures and spatial features as 2D images and this data is simultaneously utilized as a feed to our algorithm which will allow us to visualize the depth map.
\end{abstract}

\section{Introduction}
\label{sec:intro}

    Structure from Motion (SfM) is a technique used in computer vision and photogrammetry to create 3D models from a set of 2D images captured from different viewpoints. The technique involves reconstructing the 3D structure of an object or scene by analyzing the images' geometric relationships and camera parameters.
    
    The primary goal of our project is to create a robust and efficient SfM algorithm that can handle large data sets and provide accurate results in a reasonable amount of time while using a stereo camera configuration on a mobile floor robot integrated with Raspberry Pi. We are aiming at making the system autonomous.  
    
    To achieve this, all of us have conducted a comprehensive study of the existing SfM techniques, their limitations, and their strengths. SfM has several advantages over other techniques used for 3D reconstruction. It is a relatively low-cost technique that only requires a standard camera, making it accessible to a wider range of users. This technique can also handle a larger variety of object geometries and textures, making it more versatile. Additionally, SfM can be used in conjunction with other techniques, such as LiDAR and stereo vision, to create more accurate 3D models. However, SfM also has some limitations, such as the need for a large number of images to construct a reliable 3D model, and the sensitivity to camera calibration and lighting conditions. We are working towards improving the reliability of the model by also integrating it with the stereo vision technique and improving the algorithm.
    
    The project also involves the implementation of several algorithms and techniques such as feature extraction, matching, bundle adjustment, and triangulation, among others.

\section{Techniques employed}
    Our project aims to use a pair of HDR cameras in a stereo-vision configuration to reconstruct 3D models of indoor spaces. In this literature review, we will examine the existing research on the use of HDR cameras and stereo vision for 3D reconstruction, as well as related techniques such as feature detection and matching, bundle adjustment, and parallel processing.

    HDR cameras have been used in several applications related to 3D reconstruction, such as photogrammetry and computer vision. HDR imaging can improve the quality and accuracy of the reconstructed 3D models by providing more detailed and realistic images. The use of HDR cameras has been shown to be particularly useful in scenes with high-contrast lighting, where conventional cameras may struggle to capture all the detail in both bright and dark areas.

\section{Approach}
Below is the flow chart that briefly describes the process to obtain a 3D point cloud from 2D images:
    \begin{figure}[htbp]
      \centering
      \includegraphics[width=1\linewidth]{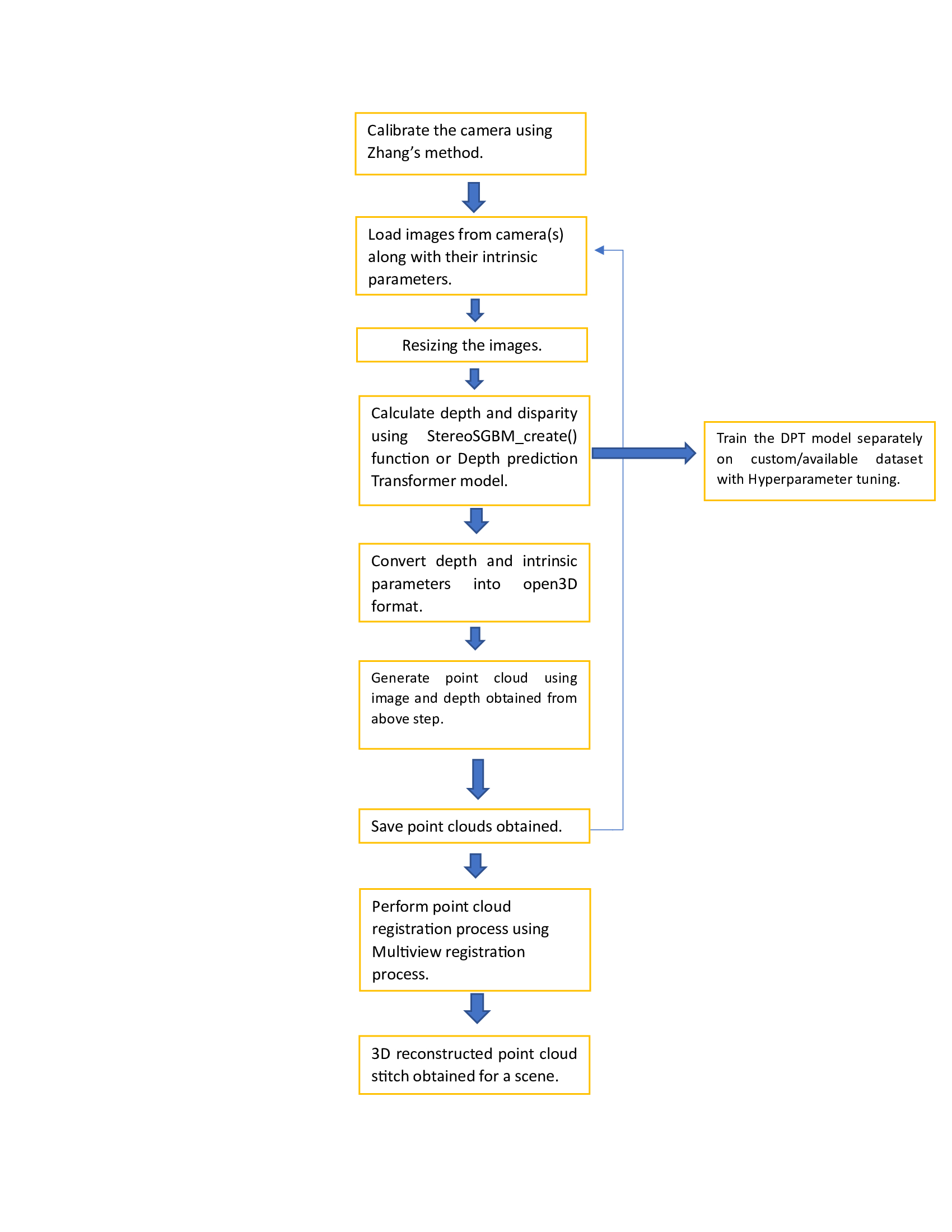}
       \caption{DPT Architecture}
    \end{figure}

\vspace{20cm}

\section{Computing Depth Map from Stereo Images}
    We, humans, have evolved to be with two eyes that we can perceive depth. And when we organize cameras analogously, it’s called Stereo-Vision. A stereo-vision system is generally made of two side-by-side cameras looking at the same scene, the following figure shows the setup of a stereo rig with an ideal configuration, aligned perfectly.

    Stereo vision is another important technique for 3D reconstruction, which involves using multiple cameras to capture images from different viewpoints. Stereo vision can provide more accurate depth information compared to other techniques such as structure from motion (SfM). Several algorithms have been developed to perform stereo matching and generate a 3D reconstruction from the stereo images.
    
    Feature detection and matching is a crucial steps in many 3D reconstruction algorithms, including the proposed project. Popular feature detection algorithms include SIFT, SURF, and ORB, while feature matching algorithms include brute force matching and RANSAC. The goal of feature detection and matching is to identify corresponding points in multiple images, which can then be used to generate a 3D point cloud.
    
    Bundle adjustment is another important technique used in 3D reconstruction to refine the camera parameters and improve the accuracy of the 3D model. Bundle adjustment algorithms typically minimize the reprojection error between the 2D image points and the corresponding 3D points. Techniques such as the Levenberg-Marquardt algorithm, Gauss-Newton optimization, and the conjugate gradient method have been used for bundle adjustment. Parallel processing can be used to speed up the 3D reconstruction process by distributing the computation across multiple processors or nodes. Parallel processing can be particularly useful for large-scale 3D reconstruction tasks that involve processing a large number of images.

    In summary, our proposed project aims to use HDR cameras in a stereo vision configuration to reconstruct 3D models of indoor spaces. This approach builds on existing research on the use of HDR imaging, stereo vision, and related techniques such as feature detection and matching, bundle adjustment, and parallel processing. The proposed project has several potential applications in areas such as robotics, interior design, and architecture, and has the potential to make significant contributions to the field of 3D reconstruction.

\section{Depth model: ViT-Hybrid}
Visual Transformers, specifically the DPT (Dense Prediction Transformers) model, are a type of deep learning architecture used for image recognition and understanding tasks. They are inspired by the success of Transformers in natural language processing and aim to apply similar principles to visual data.
The DPT model combines the strengths of convolutional neural networks (CNNs) and Transformers. CNNs are powerful in capturing local patterns and spatial hierarchies in images, while Transformers excel at capturing long-range dependencies and modeling relationships between different parts of the input.
In the DPT model, the image is divided into a grid of patches, and each patch is treated as a separate token, like how words are treated in natural language processing tasks. These image patches are then fed into a Transformer architecture, consisting of multiple layers of self-attention and feed-forward neural networks.
The self-attention mechanism allows the DPT model to capture relationships between different patches, enabling it to understand global context and capture long-range dependencies in the image. By considering interactions between all patches, the model can learn to attend to important visual features and encode them effectively.
Dilated convolutions, which are convolutional layers with increased spacing between their filter elements, are utilized in the DPT model to capture multi-scale information from the image. By incorporating dilated convolutions within the self-attention mechanism, the model can process image features at different levels of detail, effectively handling objects of various sizes.
Training the DPT model involves optimizing it with a suitable loss function, such as cross-entropy loss, and using a large, labeled dataset of images. During training, the model learns to map input images to their corresponding labels, allowing it to make predictions on new, unseen images during the inference stage.
The DPT model has shown promising results in various computer vision tasks, including image classification, object detection, and semantic segmentation. Its ability to capture long-range dependencies and understand global context makes it particularly effective in scenarios where spatial relationships between image elements are crucial.
Overall, Visual Transformers like the DPT model represent a novel approach to visual processing, leveraging the strengths of Transformers and CNNs to achieve state-of-the-art performance in image understanding tasks.
    \begin{figure}[htbp]
      \centering
      \includegraphics[width=1\linewidth]{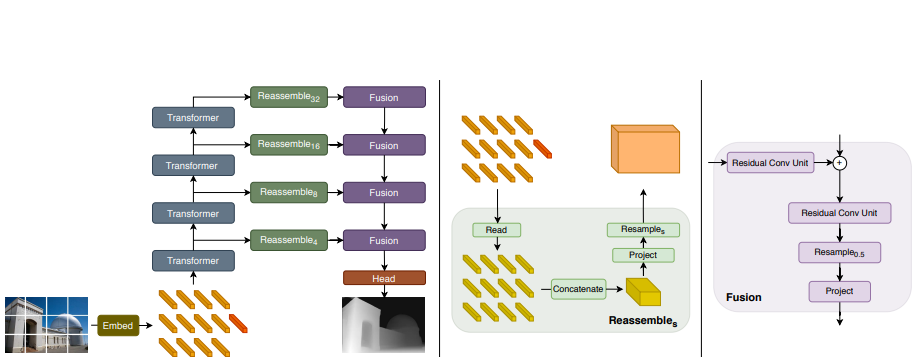}
       \caption{Flow chart}
    \end{figure}

\section{Point Cloud Registration}
Multiway registration refers to the process of aligning multiple entities or datasets simultaneously to achieve a coherent alignment. This can be applied to various types of data, such as point clouds, images, or 3D models. The goal is to find a transformation that aligns all the entities together, ensuring global consistency.

The process of multiway alignment typically involves the following steps:
\begin{enumerate}
    \item Initialization:
   - Select one entity as the reference or a common coordinate system.
   - Initialise the transformations for each entity relative to the reference.

    \item Correspondence Estimation:
       - Establish correspondences between entities to determine the relationships or associations between their elements.
       - This can be done through feature matching, nearest neighbour search, or other methods depending on the type of data.

    \item Transformation Estimation:
       - Compute the transformations (e.g., rotation, translation) that align each entity with the reference or with other entities.
       - This can be achieved through techniques such as RANSAC, Iterative Closest Point (ICP), or optimization-based approaches.

    \item Global Alignment:
       - Perform a global alignment step to refine the transformations obtained in the previous step.
       - This step aims to find a transformation that minimises the global distance or discrepancy among all entities.
       - It can involve solving an optimization problem that considers the cumulative alignment error or fitting a global transformation model.

    \item Iterative Refinement:
       - Iterate the correspondence estimation, transformation estimation, and global alignment steps to refine the alignment.
       - This iterative process helps improve the accuracy and convergence of the alignment by iteratively updating the correspondences and transformations.
\end{enumerate}

The specific techniques and algorithms used in multiway alignment depend on the type of data and the application domain. Multiway alignment finds applications in various fields, including computer vision, robotics, medical imaging, and geospatial analysis, where multiple data sources or entities need to be aligned to achieve a consistent and integrated representation.
\subsection{Iterative Closet Point (ICP)}
Point-to-plane ICP (Iterative Closest Point) is a widely used algorithm for aligning two point clouds or a point cloud to a surface model. It minimises the distance between corresponding points on the source and target surfaces.
Unlike the traditional point-to-point ICP, which minimises the distance between individual points, point-to-plane ICP considers the distance between a point and a plane. It finds the optimal transformation (translation and rotation) that minimises the sum of squared distances between the source points and the planes defined by the target points.
This approach is particularly useful when aligning point clouds with complex surfaces, as it takes into account the local surface geometry rather than just individual point positions. By aligning the source point cloud with the target surface, point-to-plane ICP can achieve more accurate registration results compared to point-to-point methods, especially in the presence of noise or partial overlap between the datasets.
 
Given the source point cloud and the target point cloud, the first step is to establish correspondences between the points in the two clouds. This can be done using a nearest neighbour search, where each point in the source cloud is matched with its closest point in the target cloud. For each correspondence (source point and target point), a plane is defined using the neighbouring points in the target cloud. The plane can be represented by a point on the plane and its associated normal vector.
 
The goal is to find the transformation (rotation and translation) that minimises the sum of squared distances between the source points and the planes defined by the target points. This is typically formulated as an optimization problem, where the objective function to minimise is the sum of squared distances.To solve the optimization problem, an iterative process is employed. In each iteration, the current transformation is refined by estimating the gradients of the objective function and updating the transformation accordingly. This process continues until a convergence criterion is met, such as reaching a maximum number of iterations or a small change in the transformation parameters.
 
Once the optimization converges, the final transformation is obtained, which aligns the source point cloud with the target point cloud based on the point-to-plane distance metric.The mathematics involved in Point-to-plane ICP can be complex, as it requires computing distances, normals, gradients, and solving an optimization problem. However, there are various algorithms and libraries available that provide implementations of Point-to-plane ICP, simplifying the usage and handling the underlying mathematical operations.
 
The goal of ICP is to align two point clouds, the old one (the existing points and normals in 3D model) and new one (new points and normals, what we want to integrate to the existing model). ICP returns rotation+translation transform between these two point clouds.
The Iterative Closest Point (ICP) minimises the objective function which is the Point to Plane Distance (PPD) between the corresponding points in two point clouds:
$$
E(\mathbf{T})=\sum_{(\mathbf{p}, \mathbf{q}) \in \mathcal{K}}\left((\mathbf{p}-\mathbf{T} \mathbf{q}) \cdot \mathbf{n}_{\mathbf{p}}\right)^2
$$
where 
np is the normal of point p has shown that the point-to-plane ICP algorithm has a faster convergence speed than the point-to-point ICP algorithm.

\section{Hardware}
Our hardware component comprises the following stack:

    \begin{figure}[htbp]
      \centering
      \includegraphics[width=1\linewidth]{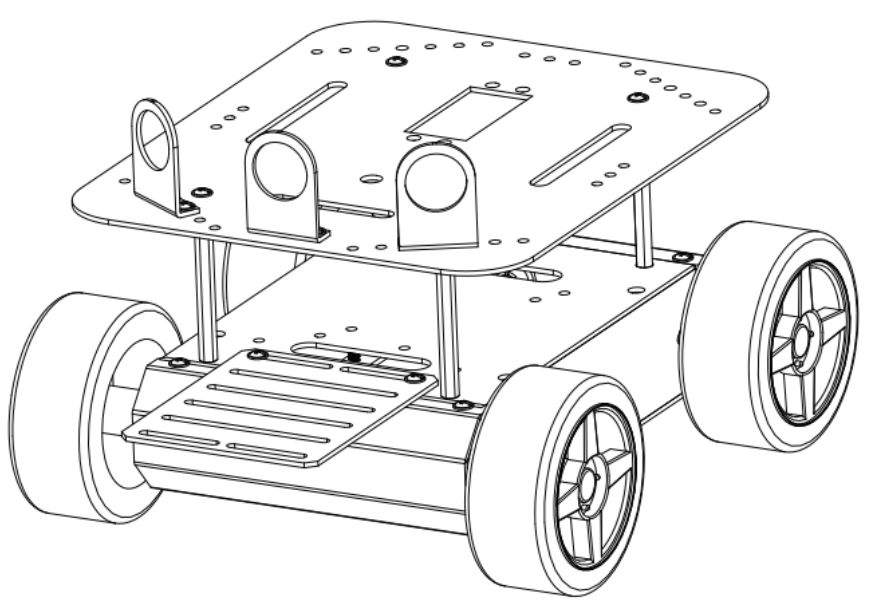}
       \caption{Robot model}
    \end{figure}
    
    The Jetson Nano is powered by a GPU and a quad-center ARM Cortex-A57 central processor, making it
    reasonable for running computationally intensive tasks like image processing and generating inferences
    from the Depth Prediction Transformer model. The Jetson Nano communicates with the Arduino Nano
    to drive the robot around within indoor space. The image processing and depth estimation pipeline on
    the Jetson Nano commonly includes the following steps:
    
    1. Image Acquisition: The Jetson Nano captures images from the 2 MIPI CSI cameras ports, where
    we have attached Raspberry Pi camera V2.
    
    2. Preprocessing: The left and right images are captured and are resized to downsample for the
    stereo depth estimation.
    
    3. Depth estimation: Depth information is calculated by two methods. First is by conventional
    disparity calculation and second is by inferencing a Dense prediction transformer to estimate
    depth using right or left camera image.
    
    4. Generate point cloud: The depth and corresponding RGB image is published to the local system
    using ROS image publisher. This data is combined by open3D functions to estimate a point cloud
    of the scene.
    
    5. Generate 3D scene of environment: The Jetson sends data to the Arduino nano using serial
    communication and the robot moves around in the space taking images from different
    viewpoints.
 \begin{figure}[htbp]
      \centering
      \includegraphics[width=1\linewidth]{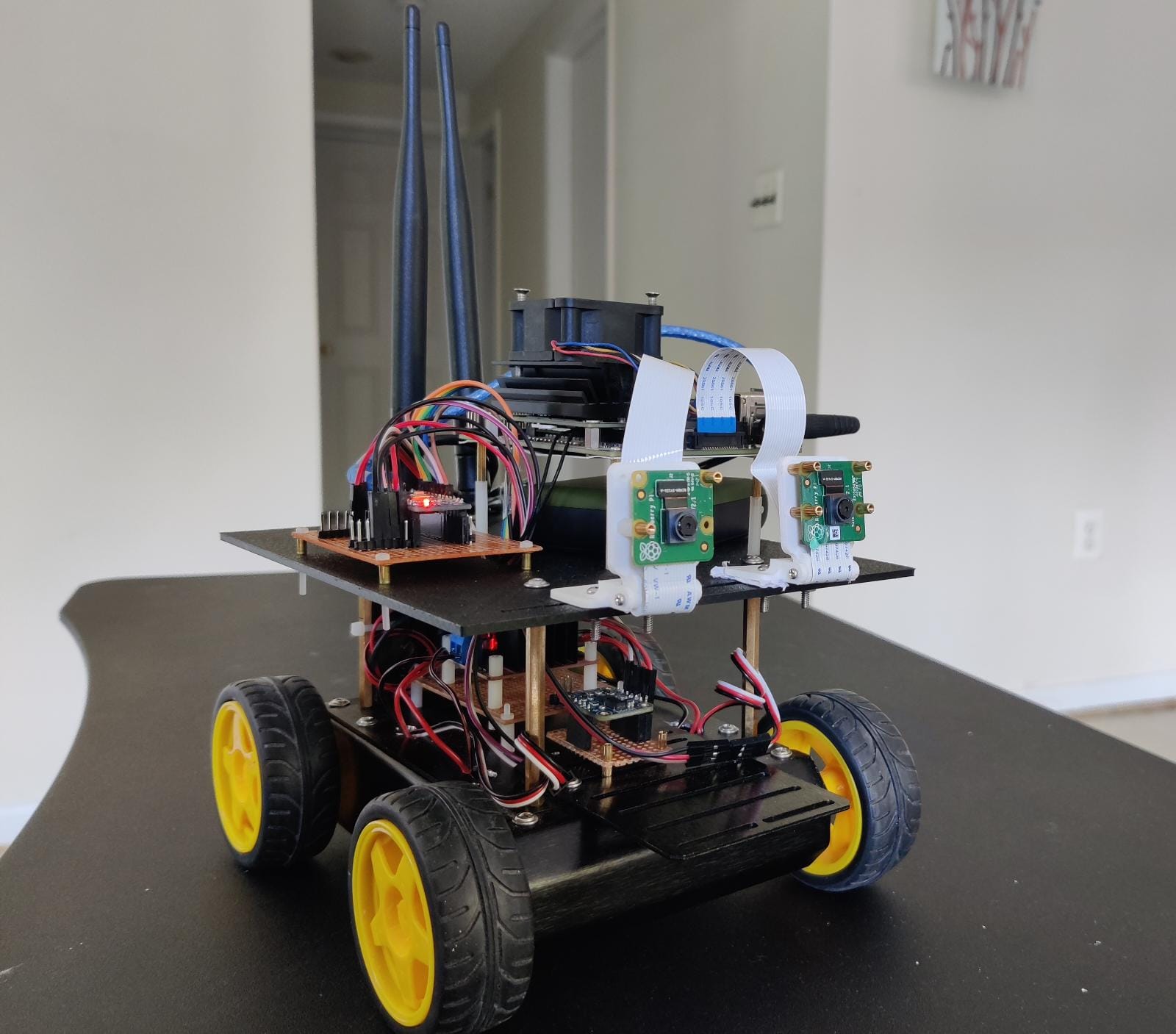}
       \caption{Robot model}
    \end{figure}
\newpage
\section{Results}
    We tested the 3D reconstruction of the indoor scene using two depth estimation techniques:
    \begin{enumerate}
        \item Conventional disparity based depth estimation using pair of stereo images. This resulted in a very coarse depth map and hence the generated point cloud does not give accurate representation of the 3D space.
        \item For the second method, we used Dense Prediction transformers(DPT) to estimate depth using a monocular camera setup. It was observed that the depth map is more robust and consistent than the disparity based method. This is shown in the results below.
        \item Another comparison is carried out between Raspberry Pi camera module and a standard Smartphone camera. It was observed that the resolution of the camera greatly affects the generated depth map and the subsequent point cloud.
    \end{enumerate}
    
    \begin{figure}[htbp]
      \centering
      \includegraphics[width=1\linewidth]{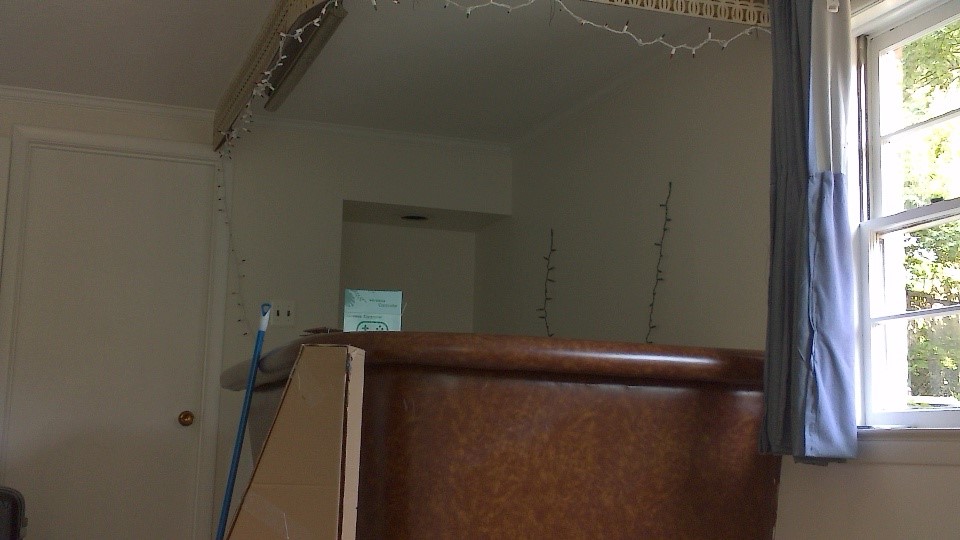}
       \caption{Raspberry Pi Left Stereo Image}
    \end{figure}

    \begin{figure}[htbp]
      \centering
      \includegraphics[width=1\linewidth]{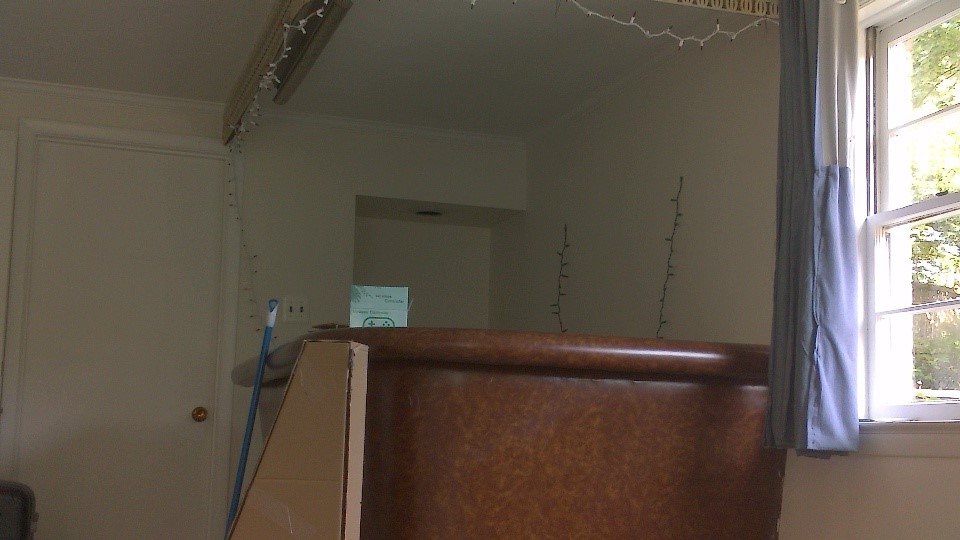}
       \caption{Raspberry Pi Right Stereo Image}
    \end{figure}
    \vspace{20cm}

    \begin{figure}[htbp]
      \centering
      \includegraphics[width=1\linewidth]{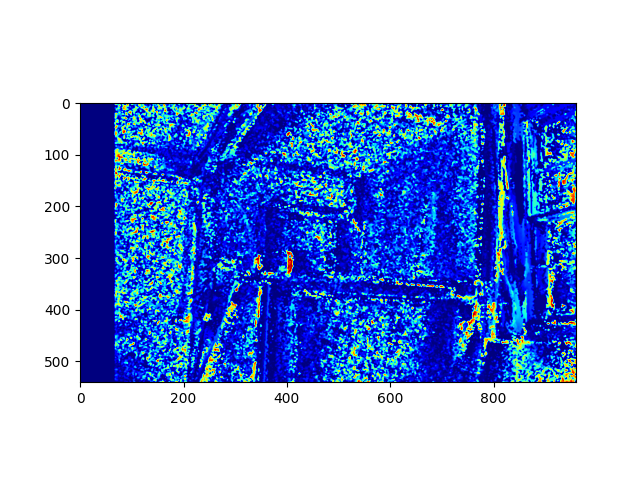}
       \caption{Depth image using Disparity Method}
    \end{figure}

    \begin{figure}[htbp]
      \centering
      \includegraphics[width=1\linewidth]{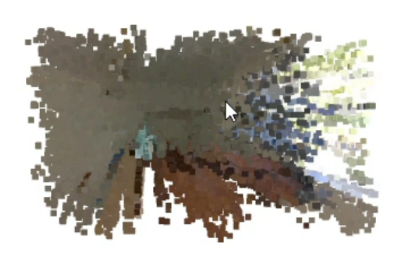}
       \caption{Disparity generated Point Cloud}
    \end{figure}

    \begin{figure}[htbp]
      \centering
      \includegraphics[width=1\linewidth]{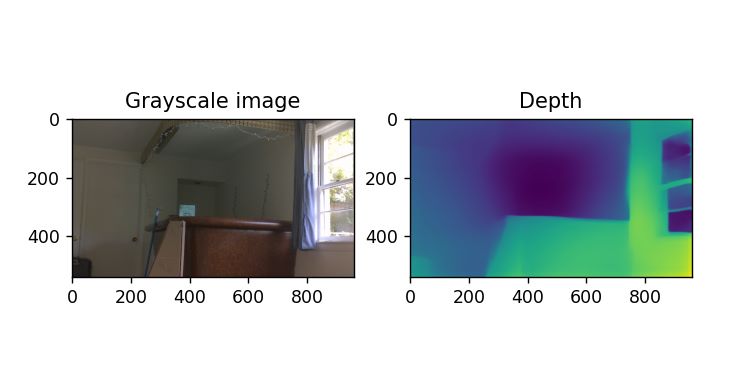}
       \caption{Raspberry Pi: Depth Estimation using DPT}
    \end{figure}
    \vspace{10cm}

    \begin{figure}[htbp]
      \centering
      \includegraphics[width=1\linewidth]{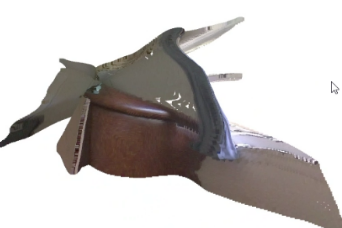}
       \caption{DPT generated Point Cloud}
   \end{figure}
   
    \begin{figure}[htbp]
      \centering
      \includegraphics[width=1\linewidth]{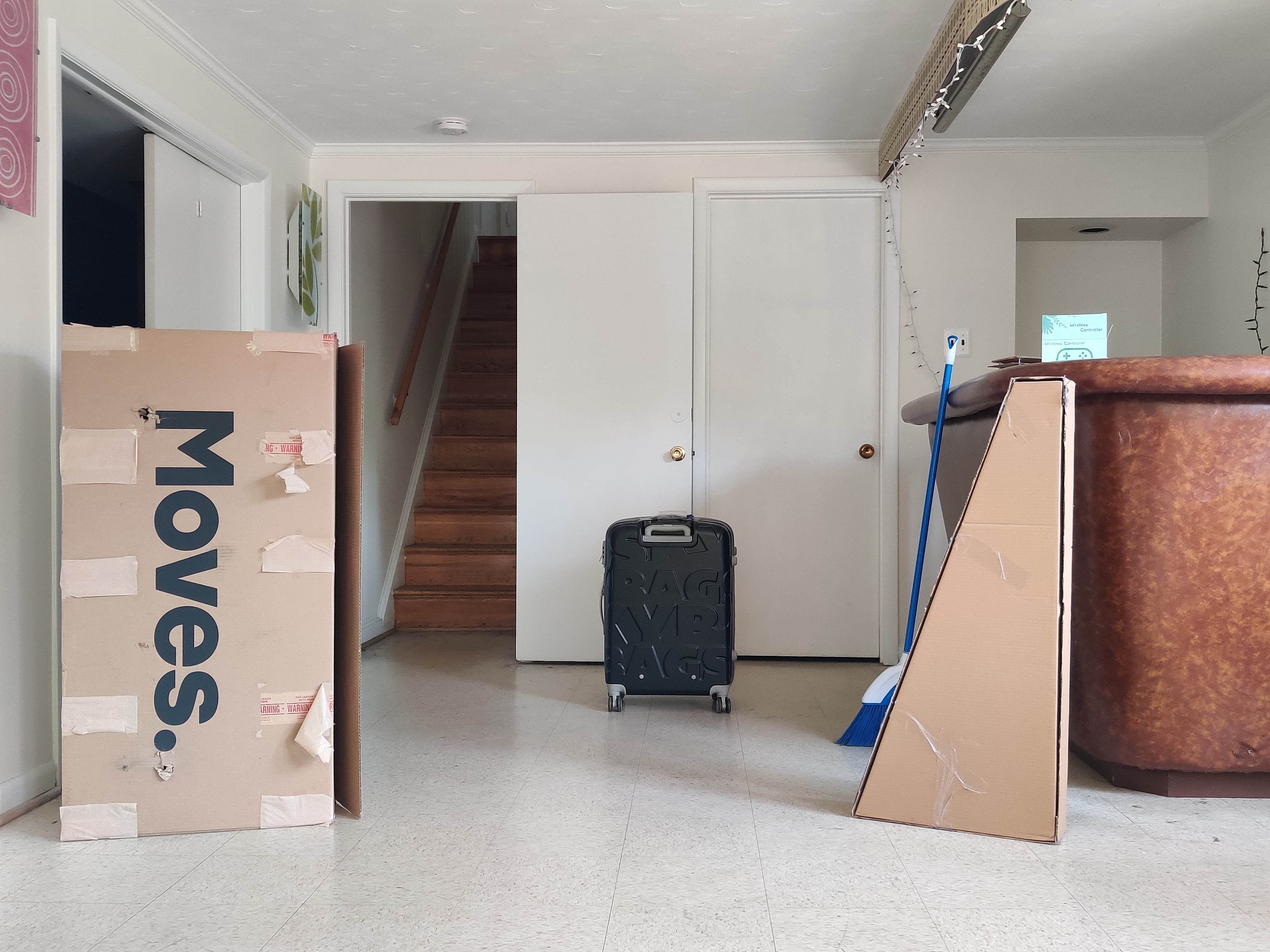}
       \caption{SmartPhone Left Setreo Image}
    \end{figure}
    \vspace{10cm}

    \begin{figure}[htbp]
      \centering
      \includegraphics[width=1\linewidth]{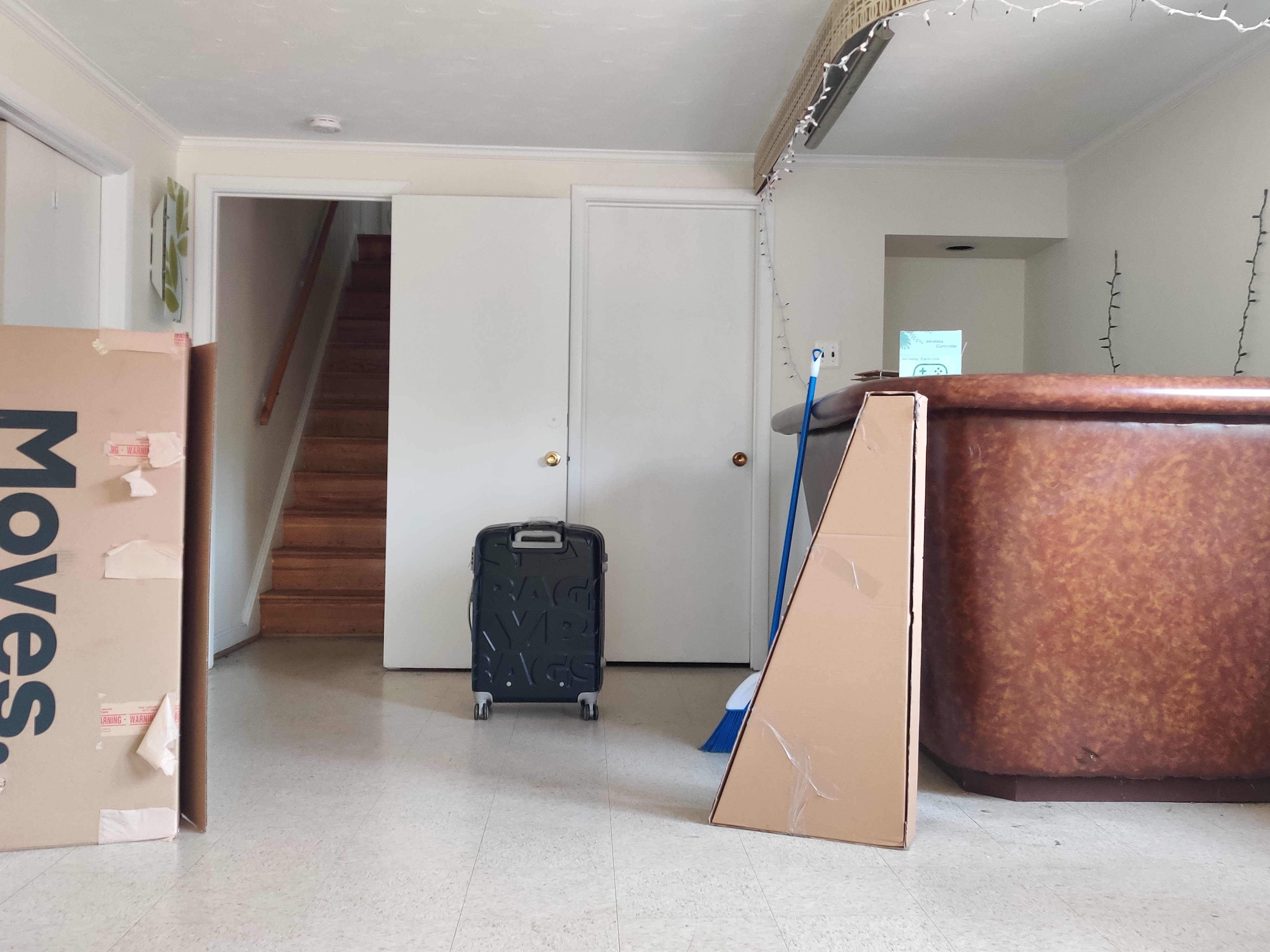}
       \caption{SmartPhone Right Stereo Image}
   \end{figure}

    \begin{figure}[htbp]
      \centering
      \includegraphics[width=1\linewidth]{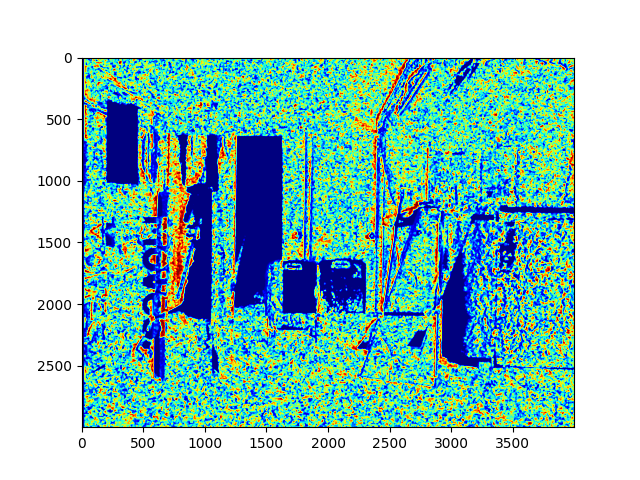}
       \caption{Depth image using Disparity Method}
   \end{figure}
   \vspace{10cm}

   \begin{figure}[htbp]
      \centering
      \includegraphics[width=1\linewidth]{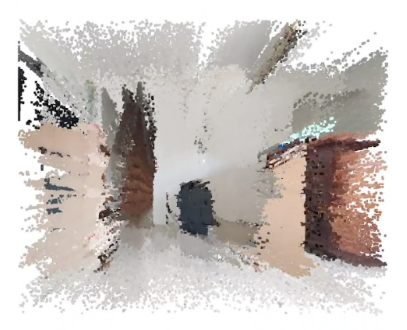}
       \caption{Disparity generated Point Cloud}
   \end{figure}

       \begin{figure}[htbp]
      \centering
      \includegraphics[width=1\linewidth]{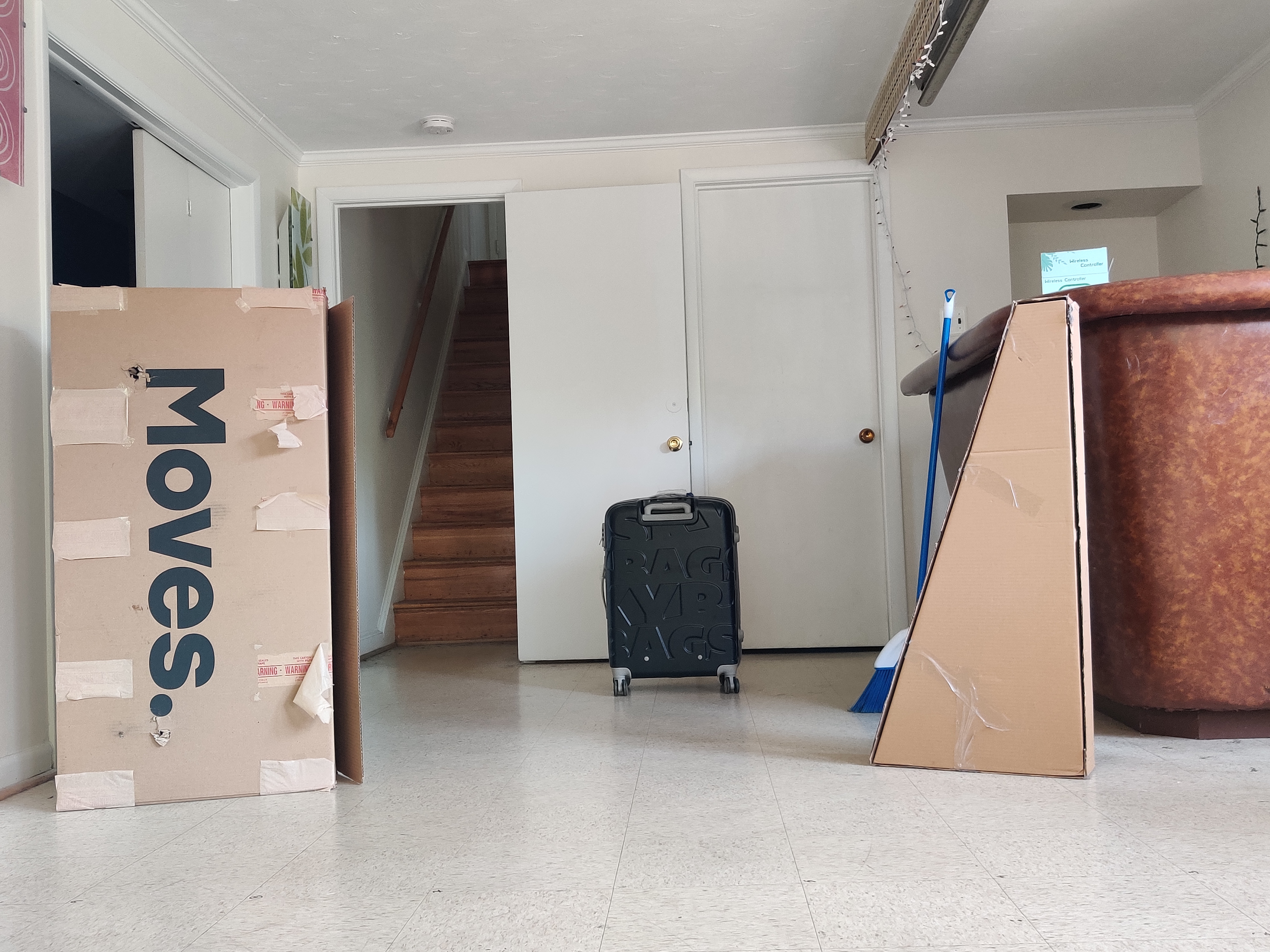}
       \caption{SmartPhone Monocular Image}
   \end{figure}
    \vspace{10cm}

    \begin{figure}[htbp]
      \centering
      \includegraphics[width=1\linewidth]{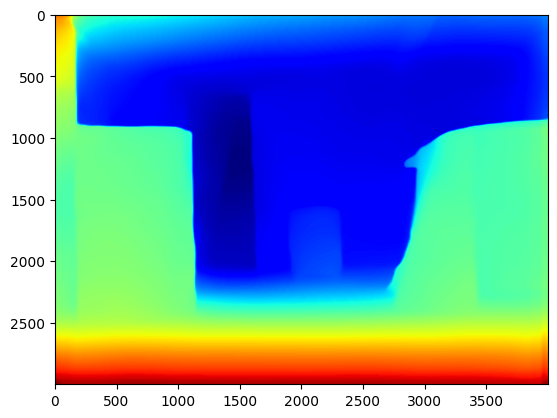}
       \caption{Depth Estimation using DPT}
   \end{figure}

    \begin{figure}[htbp]
      \centering
      \includegraphics[width=1\linewidth]{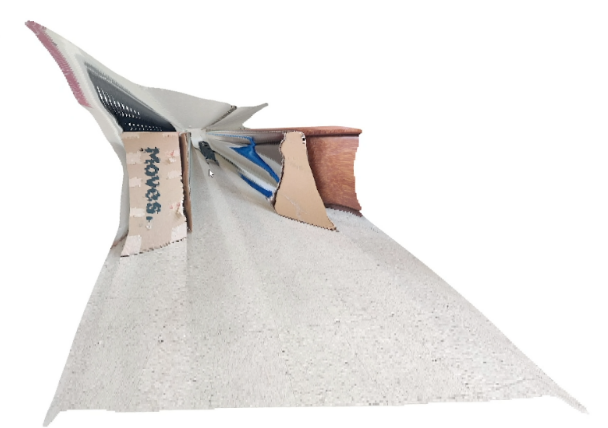}
       \caption{DPT generated Point Cloud}
   \end{figure}
       \vspace{10cm}

\vspace*{5cm}
    \begin{figure}[htbp]
      \centering
      \includegraphics[width=2\linewidth]{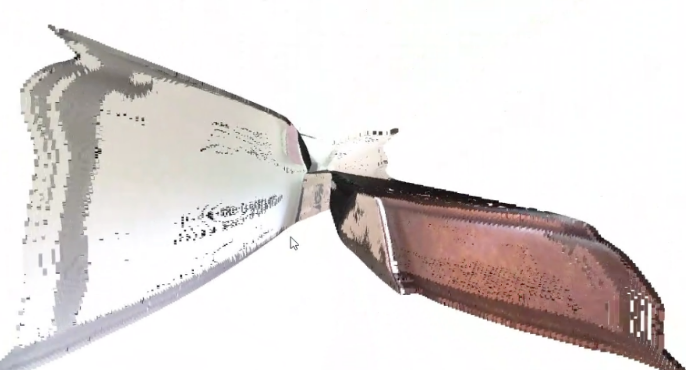}
       \caption{Registered Point Cloud}
   \end{figure}    

\clearpage
\section{Potential improvements}
\begin{enumerate}
    \item Improving the camera resolution for better quality images.
    \item Using a DL model to obtain point cloud registration can yield better results.
    \item Training the vision transformer on bigger dataset with diverse scenes and environments
\end{enumerate}

\section{References}
\begin{enumerate}
    \item https://github.com/facebookresearch/Replica-Dataset
    
    \item https://towardsdatascience.com/3-d-reconstructionwith-vision-ef0f80cbb299
    
    \item https://www.cs.cornell.edu/projects/bigsfm/
    
    \item https://www.youtube.com/watchv=DoZJaqBzSso\&abchannel=NicolaiNielsen
    
    \item http://www.open3d.org/docs/release/tutorial/geometry/pointcloud.html
    
    \item https://docs.opencv.org/3.4/d9/d0c/group calib3d.html\#ga1bc1152bd57d63bc524204f21fde6e02
    
    \item http://www.open3d.org/docs/release/tutorial/geometry/pointcloud.html

        \item https://doi.org/10.48550/arXiv.2103.13413
    
    \item https://doi.org/10.48550/arXiv.1907.01341
    
    \item http://www.open3d.org/docs/latest/tutorial/Advanced/multiway\_registration.html
    
    \item Peter Anderson, Qi Wu, Damien Teney, Jake Bruce, Mark Johnson, Niko Sunderhauf, Ian Reid, Stephen Gould, and Anton van den Hen-gel. Vision-and language navigation: Interpreting visually-grounded navigation instructions in real environments. In CVPR, 2018.
    
    \item Stanislaw Antol, Aishwarya Agrawal, Jiasen Lu, Margaret Mitchell, Dhruv Batra, C. Lawrence Zitnick, and Devi Parikh. VQA: Visual Question Answering. In ICCV, 2015.
    
    \item D. Scharstein, H. Hirschmuller, Y. Kitajima, G. Krathwohl, N. Nesic, X. Wang, and P. Westling. High-resolution stereo datasets with subpixel-accurate ground truth. In German Conference on Pattern Recognition (GCPR 2014), Munster, Germany, September 2014

\end{enumerate}
\end{document}